\documentclass[11pt]{article}
\usepackage{acl2021}  % [review]
\usepackage{times}
\usepackage[T1]{fontenc}
\usepackage[utf8]{inputenc}
\usepackage{microtype}

\usepackage{url}

\usepackage{amsfonts}
\usepackage{booktabs}
\usepackage{multirow}
\usepackage{amsmath}
\usepackage{hhline}
\usepackage{colortbl}
\usepackage{pgf-pie}
\usepackage{enumerate}
\usepackage[normalem]{ulem}
\usepackage{lipsum}  
\usepackage{inconsolata}
\usepackage{tikz,pgfplots,pgfplotstable}
\usepackage[linecolor=orange,size=scriptsize]{todonotes}
\usepackage{caption}
\usepackage{tikz}
\usepackage{pgf-pie}        
\usepackage{siunitx}   

\usepackage{ctable}

\usepackage{enumitem}

\newcommand{\wgen}{\textsc{EntDeGen}\xspace}
\newcommand{\model}{\textsc{EntDescriptor}\space}

\definecolor{MengGreen}{rgb}{0.0, 0.5, 0.0}

\newcommand{\fb}[1]{\textcolor{blue}{FB: #1}}

\DeclareMathAlphabet{\mathcal}{OMS}{cmsy}{m}{n}
\newcommand{\artw}{$\mathcal{A}_{w}$ }
\newcommand{\arth}{$\mathcal{A}_{h}$ }
\newcommand{\wsec}{$\mathcal{S}$ }
\newcommand{\ktop}{$\mathcal{K}^t$ }
\newcommand{\kfac}{$\mathcal{K}^f$ }

\newcommand{\psgs}{$\mathcal{P}$\xspace}

\usepackage{amssymb} % same
\usepackage{mathtools}
\usepackage{comment}
\usepackage{mathbbol} %{amsfonts} % Faeze added
\usepackage{subcaption}
\usepackage{arydshln}
\usepackage{fdsymbol} % for suites

\usepackage{xspace}
\pgfplotsset{compat=1.17}

\usepackage{microtype}

\title{\textit{Grounded Keys-to-Text Generation}:\\ Towards Factual Open-Ended Generation}

\author{
Faeze Brahman\textsuperscript{$\spadesuit$$\heartsuit$}\thanks{~~Work done while first author was interning at MSR.}  \hspace{.2cm} 
Baolin Peng\textsuperscript{$\diamondsuit$}  \hspace{.2cm} 
Michel Galley\textsuperscript{$\diamondsuit$}  \hspace{.2cm} \\
\bf Sudha Rao\textsuperscript{$\diamondsuit$}  \hspace{.2cm}
\bf Bill Dolan\textsuperscript{$\diamondsuit$}  \hspace{.2cm}
\bf Snigdha Chaturvedi\textsuperscript{$\clubsuit$}  \hspace{.2cm}
\bf Jianfeng Gao\textsuperscript{$\diamondsuit$} \\
\textsuperscript{$\spadesuit$}Allen Institute for Artificial Intelligence \\
\textsuperscript{$\heartsuit$}Paul G. Allen School of Computer Science \& Engineering, University of Washington \\
\textsuperscript{$\diamondsuit$}Microsoft Research, \textsuperscript{$\clubsuit$}UNC Chapel Hill \hspace{.3cm} \\
\texttt{faezeb@allenai.org} \\ %\texttt{\{baolin.peng,mgalley,sudha.rao,billdol,jfgao\}@microsoft.com},
}

\date{}

\begin{document}
\maketitle

\begin{abstract}

%\textcolor{red}{
Large pre-trained language models have recently enabled open-ended generation frameworks (e.g., prompt-to-text NLG) to tackle a variety of tasks going beyond the traditional data-to-text generation. While this framework is more general, it is under-specified and often leads to a lack of controllability restricting their real-world usage. We propose a new \textit{grounded keys-to-text} generation task: the task is to generate a factual description about an entity given a set of guiding keys, and grounding passages.
To address this task, we introduce a new dataset, called \wgen. Inspired by recent QA-based evaluation measures, we propose an automatic metric, MAFE, for factual correctness of generated descriptions. Our \textsc{EntDescriptor} model is equipped with strong rankers to fetch helpful passages and generate entity descriptions. Experimental result shows a good correlation (60.14) between our proposed metric and human judgments of factuality. Our rankers significantly improved the factual correctness of generated descriptions (15.95\% and 34.51\% relative gains in recall and precision). Finally, 
our ablation study highlights the benefit of combining \textit{keys} and \textit{groundings}.

\end{abstract}

\section{Introduction}

Converting information to text~\cite{mckeown_1985} has been a cornerstone of NLG research with the goal of improving the accessibility of knowledge to general users. % Data-to-text is often used in practical applications 
 It has found many applications such as generating sport commentaries~\cite{wiseman-etal-2017-challenges}, weather forecast~\cite{konstas-lapata-2012-unsupervised}, biographical text~\cite{lebret-etal-2016-neural}, and dialogue response generation~\cite{wen-etal-2015-semantically, wen-etal-2016-conditional}. The problem has traditionally been formulated as data-to-text generation, to generate an output given structured input such as graph, tables or key-value pairs. However, this formulation is overspecified and does not cover other \textit{open-ended} scenarios in real-world. Recent advances in large pre-trained language models (PLMs), as well as the general knowledge represented in them, have made it possible to formulate the problem as \textit{prompt-to-text} or \textit{outline-to-text}~\cite{rashkin-etal-2020-plotmachines} % or \textit{keywords-to-text}~\cite{}
generation. This offers the prospect of making NLG more broadly applicable, as such models allow input to be more parsimonious or ill-defined. However, issues such as lack of controllability and hallucination have lessened the practical applicability of this setting in real-world scenarios.

\begin{figure}[t!]
\centering
\includegraphics[width=1.0\columnwidth]{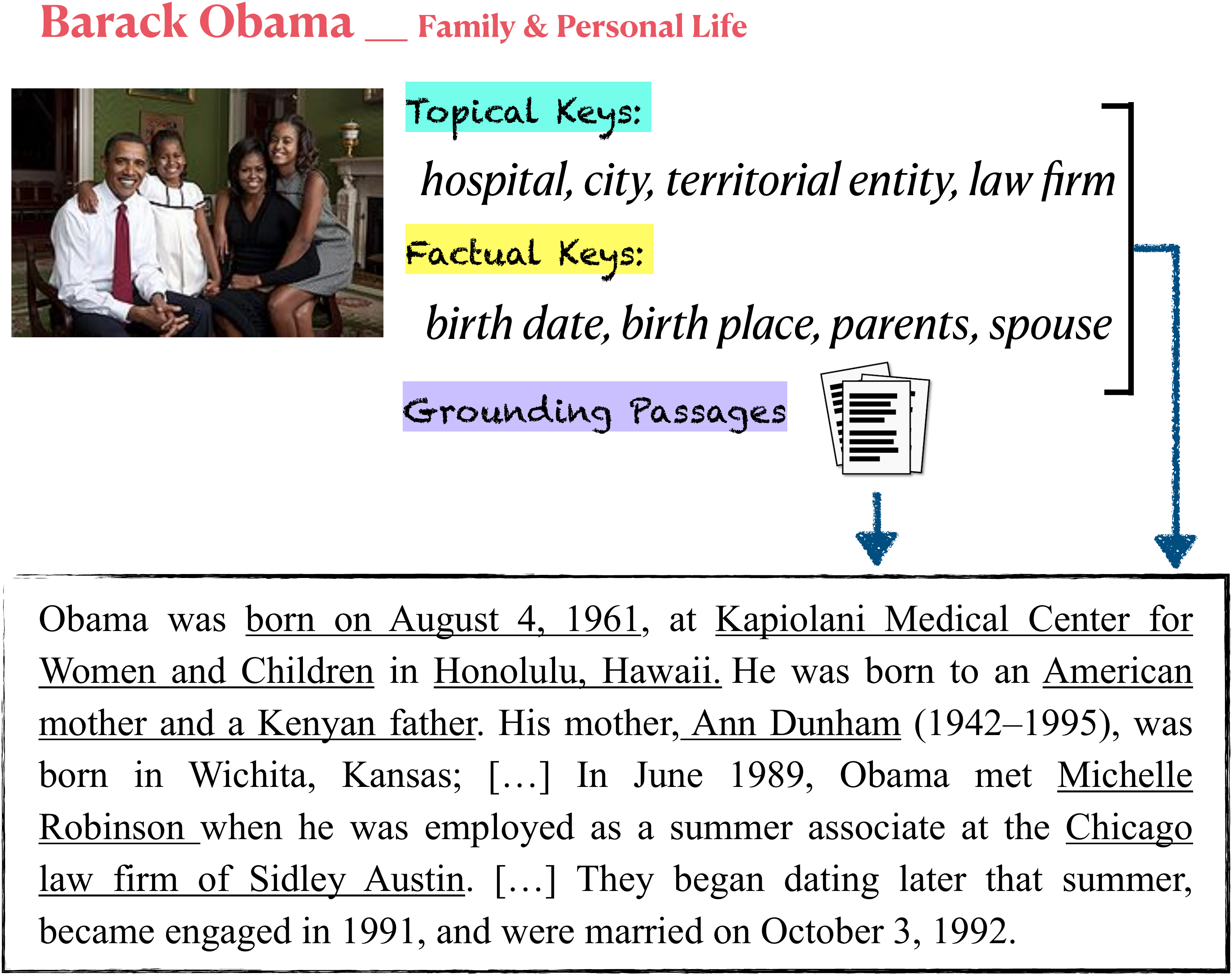}
% \vspace{-5pt}
\caption{An example from \wgen dataset. Given a set of topical and factual keys, along with multiple grounding passages, the task is to generate an entity description. Corresponding knowledge are \underline{underlined}.}
\label{fig:intro}
\end{figure}

To overcome these issues, we propose a new task, \textit{grounded keys-to-text} generation, where given a wishlist of \textit{keys} (without the values) about an entity\footnote{We consider a broad definition of ``entity'' which includes person, place, event, species, etc.} and a set of short \textit{grounding passages} as a source knowledge, the goal is to generate a factual description. An example is shown in Fig.~\ref{fig:intro}, where the task is to generate a paragraph about ``Barack Obama'', in particular about his family and personal life. Potential \textit{factual keys} in this example are ``birth date, birthplace, parents, spouse, children'', etc.
%\sout{Whereas generating about his professional career requires knowledge about \textit{occupation, offices, years in office, political party, etc.}}
The task also enables a finer-grained control over the types of entities to be included in the output via \textit{topical keys} such as ``hospital, city, law firm'' for the example in Fig.~\ref{fig:intro}. Finally, pertinent information about the entities needs to be fetched from a set of candidate \textit{grounding passages}. These passages can be obtained via internet search.
Our task differs from similar existing tasks, such as data-to-text generation~\cite{koncel-kedziorski-etal-2019-text,chen-etal-2021-wikitablet} in that, we presume keys but not values are given. This covers more open-ended scenarios in the real-world where knowledge about entities are not available in detailed structured format, is constantly changing and so have to be fetched on the fly. Moreover, this formulation offers control to the user over the generated text.

To facilitate research on grounded keys-to-text generation task, we introduce a large-scale and challenging dataset, called \wgen, with about 375K instances.
The \textit{grounded, factual} and \textit{long-form} nature of the task, brings a new challenge, %\SC{it seems to me that there are 2 challenges here which you are mixing: (1) generating coherent para-level text; (2) generating factual text. since you only address 2, why talk about 1?} 
i.e., generating paragraph-level text which is faithful to one or more grounding passages based the provided guiding keys. To address this challenge, we propose \model equipped with strong ranker %\textcolor{red}{an autoregressive ranking technique} 
to help the model focus on passages that are both relevant to the keys and complementary. We propose two rankers. Our contrastive dense ranker is based on embedding-based retrieval systems trained in a contrastive framework. Our autoregressive ranker generates a sequence of passage indices autoregressivly by modeling the probability of each passage conditioned on previously generated passages. This ranker is shown to achieve the strongest performance by modeling the joint probability of passages.

The factual aspect of generation %and the presence of groundings
also calls for a new evaluation metric. Inspired by recent fact-based evaluation for summarization, we propose an automatic metric, called MAFE, to evaluate different aspects of grounded text quality, including relevance and consistency. %, \textcolor{red}{and faithfulness}.
% We summarize our contributions as follow:
Our contributions are:\footnote{Data and code available at: \url{https://github.com/fabrahman/Grounded_Keys2Text}}
\begin{itemize}[noitemsep,topsep=0.3pt]
    \setlength\itemsep{0.3em}
    \item We introduce a new controllable entity description generation task which requires aggregating knowledge from multiple grounding passages efficiently. 
    
    \item To address this task, we also present a new dataset, called \wgen.
    
    \item We propose two ranking methods, contrastive dense and autoregressive, to select a sequence of useful passages for the model to ground in. % \SC{get rid of second sentence if needed} Our autoregressive ranker is shown to achieve the strongest performance. 

    \item We propose an evaluation metric to evaluate factual consistency in our proposed task, which highly correlates with human judgments of factuality.

\end{itemize}

\section{Task: Grounded keys-to-text Generation}

Given an entity $e$, title $t$, a set of factual \kfac$=\{k_1^f, k_2^f, ..., k_m^f\}$ and topical \ktop$=\{k_1^t, k_2^t, ..., k_m^t\}$ keys, and % a collection of % back in camera ready
grounding passages \psgs$=\{p_1, p_2, ..., p_N\}$, the goal is to generate 
a text (description) %about the entity 
%a description $h$ about the entity 
with respect to the provided keys.
\section{Dataset: \wgen}
\label{sec:wikigen}

% back in camera ready
% In this section, we describe the steps to construct the \wgen \space dataset.
% \subsection{Dataset Consruction}
% \label{sec:data-construct}

Our dataset collection strategy is based on Wikipedia and motivated by the \textsc{WikiTableT} dataset~\cite{chen-etal-2021-wikitablet}. Each Wikipedia article \artw is composed of multiple sections \wsec$=\{s_1, s_2, ..., s_n\}$. The title of the Wikipedia article is the entity $e$ whose description is to be generated and the text in each section forms a reference (gold) description, $r$. %, resulting in multiple instances in our dataset. 
For example, an article about a football player may contain sections about ``Introduction'', ``Early Life'', ``Club Career'', and ``International Career'', each forming a separate instance in our dataset. We perform the following steps for each section $s_i$ in an 
%Wikipedia 
article to obtain: factual keys, topical keys, and grounding passages. 

\noindent\textbf{Factual Keys.} \space Factual keys seek specific knowledge about an entity of interest.For obtaining factual keys, we align key-value pairs in infobox and Wikidata with each section $s_i$. For this, we took a distant-supervision approach to estimate the alignment score of each key-value pair with the section using semantic similarity and lexical precision. For semantic similarity, we compute the precision component of BERT-Score~\cite{bert-score} between the section text and the concatenation of key-value pair (\texttt{key + value}). A high value indicates that the key-value pair is semantically relevant to that section. We also measure the ROUGE-L precision score~\cite{lin2004rouge} between the section text with respect to the concatenation of key-value pair. For each instance in our dataset, we select keys whose key-value BERT-Score is greater than $0.82$, and ROUGE-L score is greater than $0.25$.\footnote{These threshold values are selected empirically from BERT-Score $\in \{0.80, 0.82, 0.84, 0.86\}$ and ROUGE-L $\in \{0.20, 0.25, 0.30\}$ based on the goodness of the alignment.}\footnote{These metrics are computed using HuggingFace dataset library: \url{https://github.com/huggingface/datasets}.}

\noindent\textbf{Topical Keys.} \space Topical keys are not tied to specific aspects of the entity of interest, but give hints on the type of other entities to be included in the output. For obtaining topical keys, we first find all hyperlinked articles \arth appearing in the section. We then use the value of the ``instance of'' or ``subclass of'' tuple in the Wikidata table of \arth as the set of topical keys for section $s_i$. For example, the hyperlinked \textit{Kapiolani Medical Center for Women and Children} in Fig.~\ref{fig:intro} is an instance of \textit{hospital} according to its Wikidata table. So it will be turned into a topical key \textit{hospital}. Both types of keys help the model to generate an output that satisfies the user’s need.

\begin{table}[t]
\scriptsize
    \centering
    \setlength\tabcolsep{6pt}
\renewcommand{\arraystretch}{1.5}
    \begin{tabular}{lccc}
    \toprule
    \textbf{} & \textbf{Train} & \textbf{Dev} & \textbf{Test} \\ 
    \midrule
    \textbf{Instances} & 267,453 & 2,500& 2,497 \\ 
    \textbf{Avg. Output Len. (token)} & 116.37 & 125.56 & 124.10  \\
    \textbf{Avg. Passage Len. (token)} & 59.98 & 60.13 & 60.25  \\
    \textbf{Avg. Top. Keys} & 4.14 & 4.15 & 4.30 \\
    \textbf{Avg. Fac. Keys} & 6.04 & 6.13 & 6.17 \\
    % \textbf{Groun. Pssg./Instance} & 40 & 40 & 40 \\
    % \textbf{COMeT} & 91.46/87.16 & 73.17/47.54 & 47.56/12.84 & 14.63/8.20 \\
    % \textbf{NLI Rationale} & 97.56/100.0 & 97.56/91.11 & 70.73/44.44 & 63.41/64.44 \\
    % \textbf{NLI Rationale w\/ highlights} & 50.00/84.62 & 38.24/53.85 & 17.65/30.77 & 17.65/30.77 \\
    % \midrule
    % \textbf{Overall} & 71.33/83.83 & 39.50/42.50 & 31.00/16.00 & 8.67/11.83 \\
    \bottomrule
    \end{tabular}
\setlength\tabcolsep{6pt}
    % \vspace{-5pt}
    \caption{\wgen  Dataset Statistics.}
    \label{tab:data-stat}
\end{table}

\noindent\textbf{Grounding Passages.} \space %Grounding passages are supposed to be short snippets of text that contain useful information about entities. 
For obtaining grounding passages, we use the documents in the WikiSum dataset~\cite{wikisum}. The documents are citations in the Wikipedia article obtained by CommonCrawl or web pages returned by Google Search. Each instance in our data has 40 grounding passages. Note that our dataset is distantly supervised, and these passages may not always contain all the facts regarding the keys. To enhance the quality of our dataset, we filter out entities for which the average Bert-Score recall of key-value pairs against the grounding passages is lower than 0.82.\footnote{Similarly, this value is chosen empirically from BERT-Score $\in \{0.80, 0.82, 0.84, 0.86\}$.}

Basic statistics of \wgen are provided in Table~\ref{tab:data-stat}.\footnote{Our dataset creation pipeline is generic and can be applied to other encyclopedic knowledge sources.} Fig. %~\ref{fig:ent_type} % back in camera ready
\ref{fig:ent_type} in Appendix depicts the diversity of \wgen \space entity domains. We associate each entity in our dataset with a domain such as Person, Place, Organization, Event, etc. 
% GPE, LOC, ORG, EVENT, COMPANY, GROUP, and MISC 
using the DBPedia knowledge-base~\cite{Lehmann2015DBpediaA}.
See Appendix~\ref{app:quality-assess} for an assessment of dataset quality.

\section{MAFE: Multi-Aspect Factuality Evaluation}
\label{sec:qa-metric}

Our proposed task is to generate a factual description. Hence, it is crucial to evaluate the factuality of the generated texts. Inspired by recent % approaches for evaluating factual consistency 
fact-based evaluation in abstractive summarization~\cite{scialom-etal-2019-answers, durmus-etal-2020-feqa, wang-etal-2020-asking}, we propose to assess the
factuality of generation through question answering (QA)% and natural language inference (NLI)
.  We evaluate factuality of a generated description $h$ with respect to % multi-modal \SC{too late to introduce new keywords like multi-modal}input 
(i) factual triples \textit{(e, k, v)} which are constructed from the entity $e$, each factual key $k$ and its value $v$,\footnote{This resembles the \textit{(s, r, o)} in the Knowledge Bases, e.g., \texttt{(Barack Obama, place of birth, Hawaii)}.} and (ii) reference (gold) description $r$%; and passages \psgs
.
In our QA based Multi-Aspect Factualy Evaluation (MAFE), questions are generated from spans in the reference and factual triples (recall), or the generated output (precision)%\SC{This sentence is not making sense to me because I don't understand the difference between reference and output.}
, and are automatically answered using the output, or reference-factual triples. %/passages
 Then, the similarity between the predicted answer and the gold answer is used to compute recall and precision.
Our evaluation framework is illustrated in Fig.~\ref{fig:eval} which accounts for both relevance (recall) and consistency (precision):
\begin{figure}[t!]
\centering
\includegraphics[width=1.02\linewidth]{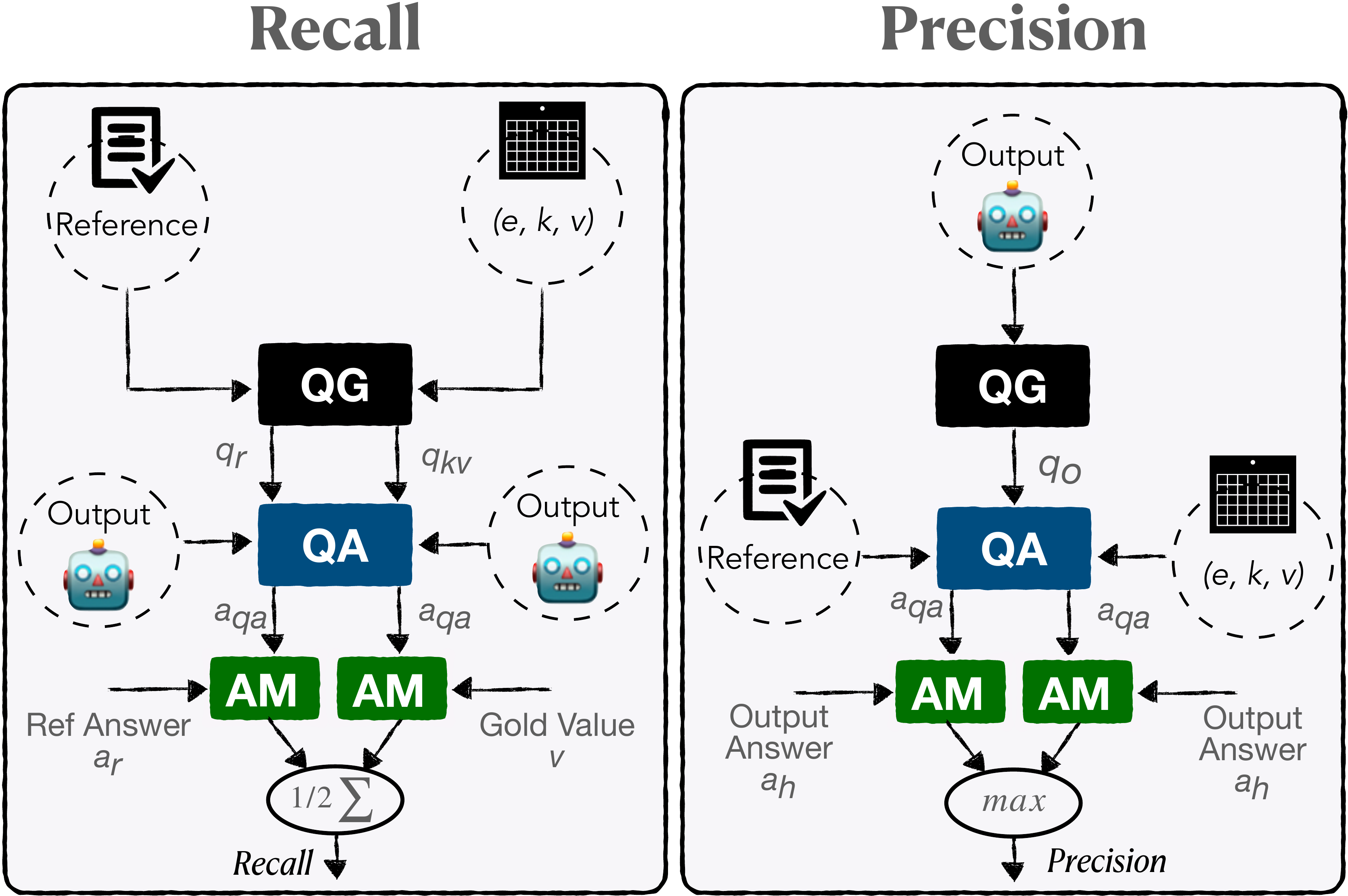} %{figures/eval_metric4.pdf}
% \vspace{-5pt}
\caption{MAFE Evaluation Framework consisting of Question Generation (QG), Question Answering (QA) and Answer Matching (AM) components. }
\label{fig:eval}
\end{figure}

\noindent \textbf{Recall} ($h \rightarrow r$) evaluates the generated output $h$ on recalling information from the factual triples \textit{(e, k, v)} AND reference $r$. For this, we generate questions that have gold answers in factual triples and reference using a Question Generation (QG) module, and obtain answers to these questions from generated output $h$ using a Question Answering (QA) module.  We define recall as the average scores of these answers when compared to the gold answers (computed by an Answer Matching (AM) module). %\sout{ from factual triples and reference}. 

% \noindent \textbf{Precision.}
\noindent \textbf{Precision} ($r \rightarrow h$) measures the amount of information contained % back in camera ready
in the generated output $h$ that is consistent with factual triples \textit{(e, k, v)} OR reference $r$. For this, we generate questions from output and obtain answers from factual triples and reference. 
We define precision as the maximum score between answers predicted from factual triples and reference.\footnote{We use \textit{maximum} because a fact contained in the model's output should either be precise w.r.t knowledge triples or the reference, but not necessarily both.}

Next, we describe the 3 modules of the evaluation framework.

\subsection{Question Generation}
Given a sentence $s$ containing an answer span $a$ (marked by special tokens), we train a QG model to generate a question $q$ (which is answerable by $a$), modeling $P_{qg}(q|s, a)$. %\textcolor{red}{The input for such a model is a sentence containing an answer span (marked by special tokens) and the output is a question which is answerable by that answer span.}
For evaluating a generated output, we gather a set of answer spans $a$ by extracting all name entities and noun phrases from each sentence $s$ (of reference or output) using spaCy\footnote{\url{https://spacy.io/}} % back in camera ready
. For generating questions from factual triples, we linearize them 
%to form a sentence 
by concatenating their constituent elements and consider the value $v$ as the answer $a$. For example, we form ``\textit{Barack Obama place of birth hawaii}'' from \texttt{(Barack Obama, place of birth, Hawaii)}. Following \citet{durmus-etal-2020-feqa}, our QG model is a BART model fine-tuned on $(s,a,q)$ triples annotated by~\citet{Demszky2018TransformingQA}.
Although the QG model is trained on natural language sentences, we found it transferring reasonably well on relational triple data because of their simple format. % back in camera ready

% \noindent \textbf{Question Answering.} \space 
\subsection{Question Answering}
Given a question $q$, and a context $c$, the QA model gives the probability of an answer $a$, modeling $P_{qa}(a|q,c)$.
For evaluating a generated output, given a question $q$ generated by the QG model from the reference and factual triples, or the output, the QA model answers it using the output, or reference and factual triples (as context $c$), respectively. For answering questions using factual triples as context, we concatenate all the linearized triples into a single 
%piece of 
text. % Leveraging recent advances in extractive QA, % bring back in camera ready
Our QA model is an ALBERT-XL model~\cite{Lan2020ALBERT:} fine-tuned on SQuAD2.0~\cite{rajpurkar-etal-2018-know}, with F1 score of 87.9\% on SQuAD2.0.
%\footnote{\textcolor{red}{Achieves F1 score of 87.9\% on SQuAD2.0.}}
%\footnote{\url{https://huggingface.co/ktrapeznikov/albert-xlarge-v2-squad-v2}} % MG: I don't think you need a url here if you already cite the work.
SQuAD2.0 support identifying \textit{unanswerable} questions, which is crucial as not all %questions can be answered using 
answers are found in a given context. % $c$.

% \noindent \textbf{Answer Matching.} \space 
\subsection{Answer Matching}
The common approach to assess the answers given by a QA model (compared to gold answers) is to use F1-score, which is based on exact matching of \textit{n}-grams. We argue that is problematic in our case when correct answers are lexically different. For example, Sport:\textit{``professional wrestling''} can be realized as \textit{``She is a wrestler [...]''}. The F1-score does not capture these lexically varied but correct answers. Therefore, we propose using an NLI model to compare the similarity of two answers. Given the generated question $q$ from the reference, to compare the reference (gold) answer and the predicted answer, we concatenate each answer with the question separately to form the premise and hypothesis for the NLI model. For example, for the question \textit{``What sport did Mr. Kenny Jay play?''}, we pass the following to the NLI model:

\vspace{5pt}

\noindent\fbox{
\parbox{0.95\linewidth}{
%\parbox{\linewidth - 2\fboxsep}{
%\noindent 
\texttt{Premise:}\textit{What sport did Mr. Kenny Jay play? professional wrestling} \\
%\noindent 
\texttt{Hypothesis:}\textit{What sport did Mr. Kenny Jay play? wrestler }
% premise and as the hypothesis 
}
}

\vspace{5pt}

\noindent We give the predicted answer a score of $1$ if the NLI model predicts entailment, and a score of $0$ if it predicts contradiction. 
%In the case of
For
neutral%\SC{need a short justification on why special processing for neutral}
, we compute the BERTScore~\cite{bert-score} comparing the contextualized representations of the two answers. For the NLI model, we use RoBERTa~\cite{DBLP:journals/corr/abs-1907-11692} fine-tuned on MNLI~\cite{mnli}, 
with an accuracy of 90\% on MNLI. We included examples of comparison between NLI and F1 score in Table~\ref{tab:app:f1-nli}.

\begin{comment}
\noindent \textbf{Extrinsic Precision} evaluates the factual consistency (lack of hallucination) by considering any questions (generated from the model's output) with similar answers in the grounding passages \psgs as factual and hallucinated otherwise.
\end{comment}

\begin{figure*}[t!]
\centering
\includegraphics[width=1.0\linewidth]{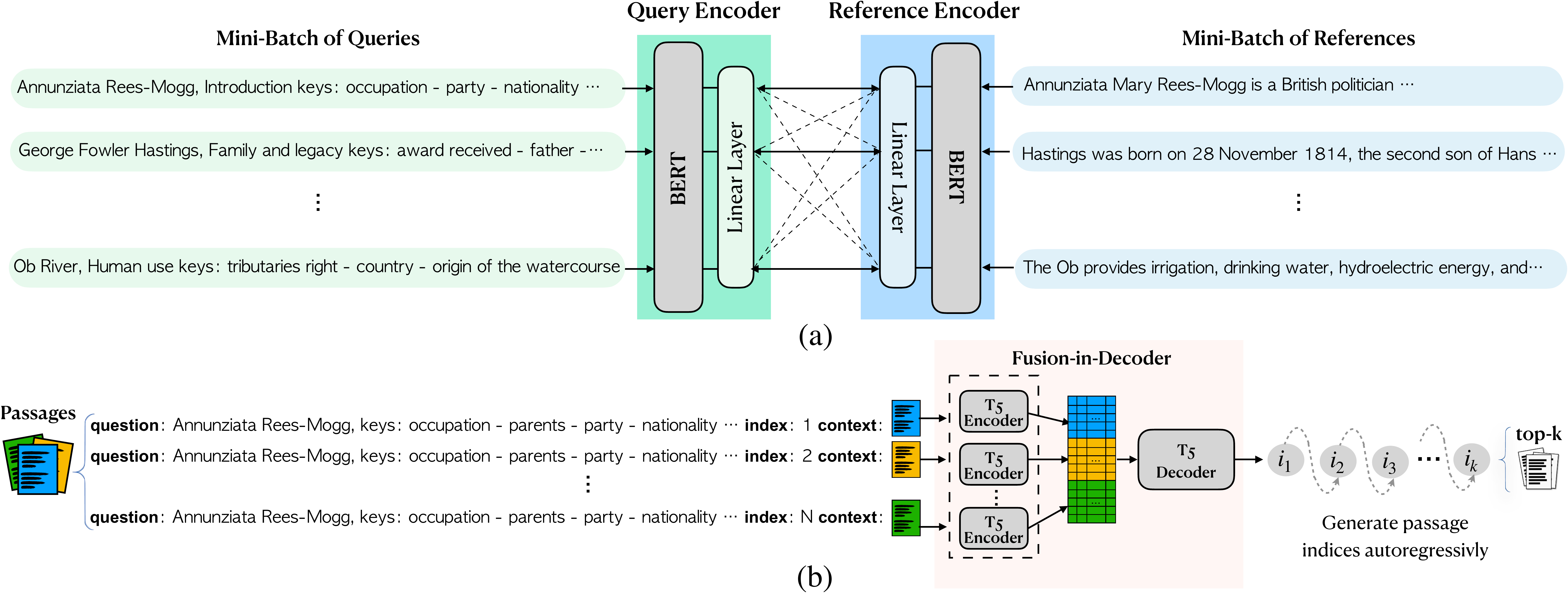}
% \vspace{-5pt}
\caption{Two proposed passage rankers: (a) Contrastive Dense (b) Autoregressive.}
\label{fig:rankers}
\end{figure*}

\section{\model}
\label{sec:model}

%Given an entity and a set of keys, % , and a collection of grounding passages
The \model model needs to fetch relevant passages on the fly to generate a factual description. %about the entity of interest. 
For this, we equip our \model model with a \textit{Passage Ranker} (\S\ref{sec:ranker}). Given the entity, keys and a set of ranked passages, the \textit{Descriptor Generator} (\S\ref{sec:baselines}) then generates an entity description. 
\begin{comment}
We equip our Entity-Descriptor model with a ranking component described below to fetch relevant grounding passages on-the-fly given an entity of interest and a set of keys.
\end{comment}

\subsection{Passage Ranker}
\label{sec:ranker}

Each instance in our dataset is accompanied by a set of candidate grounding passages. However, not all passages contain useful knowledge about certain aspects of an entity, i.e., the provided \textit{factual} and \textit{topical keys}. We, therefore, introduce a ranking stage where we rank the grounding passages \psgs given the entity, title, and a set of keys as query $q$. The ranker outputs top-\textit{k} passages %. We then only use the $\{p1,...,p_k\} \subset $ \psgs as input to the models. 
$\{p_1,...,p_k\}\subset $ \psgs{}, which are then used to ground the Descriptor Generator. % as input to the EntityDescriptor. 
Below, we describe two baseline rankers, namely ROUGE-2 and tf-idf rankers, and two proposed rankers, namely contrastive dense and autoregressive rankers.% \fb{How to say the latter two is our main proposed approaches and the first two are just baselines?}

\noindent\textbf{ROUGE-2 (\textit{oracle}).}\space This ranker ranks passages according to their ROUGE-2 recall against the reference. % We consider this % bring back in camera ready
This is akin to \textit{oracle} ranking as we use information in the reference to do the ranking.

\noindent\textbf{Tf-idf.}\space This ranker ranks passages using their tf-idf score following~\citet{wikisum}.

\noindent\textbf{Contrastive Dense. }\space This ranker learns and then compares dense representations of queries and passages using contrastive training. 
% Drawing inspiration from recent advances in embedding-based retrieval systems such as REALM~\cite{guu2020realm}, and DPR~\cite{KarpukhinOMLWEC20},
We train a dense ranker (shown in Fig.~\ref{fig:rankers}(a)) which is inspired by recent embedding-based retrieval systems such as REALM~\cite{guu2020realm}, and DPR~\cite{KarpukhinOMLWEC20}. We follow a distant supervision approach~\cite{jernite20} by using the reference descriptions $r$ instead of the gold passages as supervision signal. %\sout{This is mainly intuitive as the Wikipedia paragraphs are stylistically comparable to the grounding passages}
For each instance, we form a query $q_i$ by concatenating the entity, title and the set of keys. We thus construct a dataset of $(q_i, r_i)$ pairs and use a bi-encoder architecture to project queries and references to 128-$d$ embedding space. We use a contrastive framework with in-batch negatives where the idea is to push encoded vector of a query closer to its corresponding reference vector, but away from other reference vectors in the batch. Formally, we optimize the following Cross-Entropy loss with in-batch negatives:
\begin{equation}
     \mathcal{L} = - \sum_{(q_i, r_i)\in mB} \log \frac{\exp(\mathbf{q_i}. \mathbf{r_i})}{\sum_{r_j\in mB} \exp (\mathbf{q_i}. \mathbf{r_j})}
\end{equation}

\noindent where $\mathbf{q_i}$ and $\mathbf{r_i}$ are encoded query and reference vectors, and $mB$ denotes the mini-Batch. We use mini-batches of 1024, and initialize the encoders with distilled-BERT~\cite{distill-bert, devlin-etal-2019-bert}. Two projection layers are then learned for queries and references. Once the ranker is trained, % we use the reference encoder to score each grounding passage relative to the input q
we use the reference encoder to encode each grounding passage $p_i$ and score them based on their dot product similarity w.r.t vector representation of query $\mathbf{q_i}$. We then use the top-$k$ passages as input to Descriptor Generator.

\noindent\textbf{Autoregressive.}\space
In the previous ranker, passages are scored independently according to their relevance to the input query $q$. However, an ideal ranker should % be optimized to retrieve % back in camera ready
select relevant yet diverse passages. To achieve this goal, we develop an autoregressive ranker with an encoder-decoder architecture (shown in Fig.~\ref{fig:rankers}(b)) where the encoder process the entire set of passages \psgs, and the decoder generates a sequence of $k$ passage indices. The autoregressive nature enables modeling the joint probability of passages $P(p_1, ..., p_N| % e, t,$ \kfac , \ktop$)$
q)$. Similar text-to-index framework showed promising results for \textit{sentence ordering}~\cite{basu-roy-chowdhury-etal-2021-everything} and \textit{multi-answer retrieval}~\cite{min-etal-2021-joint}.
To enable encoding the entire set of passages (in our case 40), we use the Fusion-in-Decoder (FiD) architecture following~\citet{izacard2020leveraging}.

The FiD architecture takes the input query %(formed similarly as in contrastive ranker)
(concatenation of entity, title and set of keys) as well as each individual passage independently as inputs to its encoders. %More precisely, t
The query is concatenated with each passage and its positional index % and is independently fed to the encoder.
using special tokens: \texttt{question:} \texttt{[Entity]} $e$ \texttt{[Title]} $t$ \texttt{[Keys]} \kfac + \ktop \texttt{index:} $i$ \texttt{context:} $p_i$. % The output of encoder for each individual passage is $\textbf{h_i} \in \mathbb{R}^{L\times d}$
The encoders output 
%hidden  (clear from context)
representations $\mathbf{h_i}$ $\in \mathbb{R}^{L \times d}$ for each individual passage $p_i$, where $L$, and $d$ are input length and hidden dimension, respectively. The concatenation of encoders' representations $\mathbf{H} = [\mathbf{h_1};...;\mathbf{h_N}] \in \mathbb{R}^{L.N \times d}$ is passed to decoder which in turn generates a sequence of passage indices $\{i_1, ...i_k\}$. %A similar text-to-index framework has shown promising results for \textit{sentence ordering}~\cite{basu-roy-chowdhury-etal-2021-everything}. 
The corresponding  top-\textit{k} passages $\{p_{i_1},...,p_{i_k}\}$ are then used to ground the Descriptor Generator. % entity description generation task. 
All encoders and the decoder are initialized with %pre-trained 
T5~\cite{2020t5}. This ranker is trained %on a training dataset in which % back in camera ready
using the \textit{silver} sequence of passage indices obtained by ROUGE-2 \textit{(oracle)} ranker. % the \textit{oracle} method, i.e., ROUGE-2.

\subsection{Descriptor Generator} %{Transformer-based Baselines}
\label{sec:baselines}

\noindent\textbf{Extractive.}\space We build an extractive baseline using QG and QA models. For this, we convert an entity name and each factual key in our input into a natural language question using a seq2seq model. We then use a strong extractive QA model, namely ALBERT-XL fine-tuned on SQuAD2.0, to answer these questions using all grounding passages as the context
(Each grounding passage is passed separately).
%\footnote{Each grounding passage is passed separately.}
Finally, we concatenate all sentences from the groundings which contained the most confident answers as our final output.

\noindent\textbf{Abstractive.}\space  
We build strong abstractive baselines by fine-tuning several transformer-based PLMs. This include %standard left-to-right LMs, namely GPT2-M~\cite{}\fb{Thinking of removing GPT2 baseline?}, and 
encoder-decoder models, namely \textbf{BART-Large}~\cite{lewis-etal-2020-bart}, \textbf{T5-large}~\cite{2020t5}, and \textbf{PEGASUS}~\cite{zhang2019pegasus}.  All baselines take inputs in the form of \texttt{[Entity]} $e$ \texttt{[Title]} $t$ \texttt{[Keys]} \kfac + \ktop \texttt{[docs]} $P^{k}$ and generate the entity description. 
Note that during training our generation models, we use the top-$10$ grounding passages obtained by \textit{oracle} ROUGE-2 ranker.\footnote{Using the passages obtained from other rankers during training degrades the performance.}
See Appendix~\ref{app:imp_detail} for 
%implementation 
details.
%Implementation details %of experiments
%are provided in Appendix~\ref{app:imp_detail}.

\section{Experiments}

\begin{comment}
\subsection{Implementation Details} 
We use the Transformer library~\cite{Wolf2019HuggingFacesTS}. Each baseline was trained for 3 epochs with effective batch size of 8, and initial learning rate of 5e-6 for T5 and BART, and 1e-4 for PEGASUS. We use the maximum input length of 512 tokens. During inference, we use beam search decoding with 5 beams, and repetition penalty of 1.2. 

The contrastive dense ranker was trained for 10 epochs with 2e-4 learning rate. The autoregressive ranker was trained for total 30,000 steps with learning rate and weight decay of 1e-5 and 0.01, respectively.
\fb{decoding for QG?/ k in top-k passages/ BART-large xsum, pegasus xsum}
\end{comment}

\subsection{Automatic Evaluation}
\label{subsec:auto-eval}
% Following previous works
Beside our proposed MAFE metric, we use several widely used automatic metrics like \textbf{BLEU}~\cite{papineni2002bleu}, % that measures overlap of \textit{n}-gram up to $n=4$ % back in camera ready
%\textbf{ROUGE-\textit{n}} ($n{=}1,2$),
and \textbf{ROUGE-L}~\cite{lin2004rouge}. However, recent works~\cite{dhingra-etal-2019-handling} have raised concerns on the usage of these metrics for automatically constructed data-to-text dataset as they fail to consider divergent reference texts. % bring back in camera ready
We also use \textbf{PARENT}~\cite{dhingra-etal-2019-handling}\footnote{We use the co-occurrence version % as it showed higher correlation with human evaluation % back in camer ready
which is recommended when paraphrasing is involved between data and text. } that considers similarity of generation to both data (in our case factual triples) %\footnote{We assume values are available during evaluation.} % back in camera ready
and the reference. Lastly, we use \textbf{BERTScore}~\cite{bert-score} which computes alignment between % tokens of reference and generated output using BERT representations. % use this in camera ready
BERT representations of reference and generated output.

\begin{table}[t]
\footnotesize
    \centering
    \setlength\tabcolsep{5.5pt}
\renewcommand{\arraystretch}{1.5}
\begin{tabular}{lccccc}
\toprule
Metric & BLEU & R-L  & PAR  & BERT-S & MAFE \\ \hline
Corr.  & 48.00 & 56.75 & 40.80 & 56.76   & \textbf{60.14} \\\toprule
\end{tabular}

    % \vspace{-5pt}
    \caption{Paragraph-level Pearson correlation coefficient between automatic metrics and human judgement of factuality. All correlations are significant at $p< 0.001$.}
    \label{tab:corr}
\end{table}

\subsection{Evaluation of MAFE}
\label{sec:mafe-eval}
We propose a metric for evaluating factuality, MAFE. To evaluate MAFE as a metric, we compute its correlation with human judgments of factuality. We take a random set of 297 BART-L generated outputs using different rankers. The instances include diverse set of entity domains (see Fig.~\ref{fig:ent_type_corr} in Appendix). We collect human judgments of factuality on this subset using Amazon Mechanical Turk (AMT). Three annotators judged the recall-oriented and precision-oriented factuality of each generated paragraph. For evaluating recall-oriented factuality, we present each sentence of the reference one at a time and ask annotators how well the sentence is supported by the content in the generated paragraph. The annotators have to choose from a Likert scale of 1-5 (1 being very badly supported, 5 being very well supported).\footnote{We find the Likert scale to be more suitable than binary decision because each sentence might contain multiple facts.} We also present each factual triple one at a time and ask annotators if it is supported by the content in the generated paragraph.\footnote{Factual triples are presented in the form of \textit{e (k; v)}. E.g. \textit{Henry Stanton (placeofburial; West Point)} which is read as ``The place of burial of Henry Stanton is West Point.''} % back in camera ready
 For evaluating precision-oriented factuality, we switch references and factual triples with generated paragraphs, i.e., we show the generated paragraphs one sentence at a time and ask how well the sentence is supported by the reference and all factual triples. We then average scores across all sentences. See Appendix~\ref{supp:human-design}
for details and screenshots of annotation layout. 
% The annotation pipeline is fully described in the \textcolor{blue}{Appendix~\ref{}}.

To account for recall and precision oriented values, we measure correlations between human judgment F1 ($2\frac{rec.prec}{rec+prec}$) with MAFE-F1 % as well as all % bring back in camera ready
and other automatic metrics. 
% The results are reported in Table~\ref{tab:corr}. % back in camera ready
According to Table~\ref{tab:corr}, MAFE shows a higher correlation with the human judgment than other metrics. Hence, we include MAFE in our experiments to gauge the factuality of generations.

\subsection{Results}
\label{sec:results}
\noindent\textbf{Performance of Different Baselines.} \space Table~\ref{tab:auto-res-all} reports the performance of different baselines for the task of entity description generation. According to the results, Extractive performs poorly compared to other abstractive baselines. This is mainly because it lacks the narrative flow required for a coherent output. Comparing all abstractive baselines, when they are given \textit{oracle} groundings (defined in \S\ref{sec:ranker}), shows that BART outperforms T5 and PEGASUS in general on all $n$-gram overlap-based, PARENT, as well as BERTScore metrics.
%\footnote{
Uni/bi-gram overlap (R-1,R-2) are reported in Table~\ref{tab:rouge-all}.
%}

When comparing baselines with respect to factuality using our MAFE metric, we see that  BART in general generates paragraphs that are significantly more consistent (precise) with respect to factual triples and reference. Whereas, T5 is slightly better at content-selection (measured by recall). 

%\begin{table*}[t]
\begin{table}[t]
\scriptsize
    \centering
    %%% correct precision
\setlength\tabcolsep{4.2pt}
\renewcommand{\arraystretch}{1.5}
%\begin{tabular}{l|l|cc|c|c|cc}
%\begin{tabular}{l|cc|c|c|cc}
\begin{tabular}{l|cccc|cc}
\toprule
\multirow{2}{*}{\textbf{Model} +Ranker} & \multirow{2}{*}{\textbf{BLEU}} & \multirow{2}{*}{\textbf{R-L}} & \multirow{2}{*}{\textbf{PAR}} & \multirow{2}{*}{\textbf{BERTS}} & \multicolumn{2}{c}{\textbf{MAFE}}               \\ \cline{6-7} 
 & & & & & R & P\\ \hline
\textbf{Extractive}             &    2.94      &    19.85     &   14.22  & 82.44 & 19.01 & 18.91  \\ \hline
\textbf{T5-Large} &        5.41     &     15.09   &    20.58  & 84.25  &  17.65 &  17.38 \\
+Tf-idf    &          5.50    &  26.70       &   21.42  & 84.53 & 17.68 & 18.35  \\
+Contrastive Dense             &         6.89   &  28.14  &     22.44 & 84.92   &  19.00 &   19.66 \\
+Autoregressive          &     7.24  &   28.64      &   23.45   & 84.96 & \textbf{19.48}    &    19.97    \\ \cdashline{1-7} 
+\textit{Rouge2 (oracle)} &         8.56    &  30.84   &     25.97  & 85.41   &    20.99   &   22.01   \\ \hline
\textbf{BART-L}              &        7.03   &  30.82    & 23.57   & 86.10  &  17.43 &   23.13  \\
+Tf-idf                  &         6.97   &   31.14    &    23.96  & 86.23   & 17.45 &    24.02    \\
+Contrastive Dense             &       8.25  &   32.46    &    24.67  & 86.48    &   18.55 & 26.00  \\
+Autoregressive          &       \textbf{8.69}   &    \textbf{32.97}    &   \textbf{25.40}  & \textbf{86.58}  &    19.11  & \textbf{26.71}   \\ \cdashline{1-7} 
+\textit{Rouge2 (oracle)} &       9.82    &  34.87    &     27.25   & 87.01   &     20.28    & 29.32  \\ \hline
\textbf{PEGASUS}                &           6.49  &   27.11         &    22.88  & 83.65     & 15.34 &  22.72 \\
+Tf-idf                  &         6.34   &       27.25     &         22.68    &  83.74  & 14.79 &  23.72 \\
+Contrastive Dense     &         7.99   &   28.94     &     24.10  & 84.43 & 16.40 &  24.70  \\
+Autoregressive          &       8.55   &   29.75     &  25.38  & 84.54 &    17.16  &  25.34   \\ \cdashline{1-7} 
+\textit{Rouge2 (oracle)} &           10.05   &    31.71   &    27.72  & 84.97  &     18.07  &   26.73    \\ \bottomrule
\end{tabular}
\setlength\tabcolsep{8pt}
    %\input{tables/all_auto_res_2}
    % \vspace{-5pt}
    \caption{BLEU, ROUGE-L, PARENT, BERTScore, and MAFE scores for different unranked models,  as well with adding different rankers. Models consistently perform better when using autoregressive ranker.}
    \label{tab:auto-res-all}
%\end{table*}
\end{table}

\noindent\textbf{Performance of Different Rankers.} \space We now investigate the effect of different rankers on generation performance. %To this end, w % back in camera ready
For this, we compare baselines using different rankers (see Table~\ref{tab:auto-res-all}). All models perform better when they are given top-$k$ ranked groundings than their \textit{Unranked} baselines. For all generation models, the proposed contrastive and autoregressive rankers significantly outperform the tf-idf baseline ranker. This is because tf-idf ranker only finds passages that feature sparse words from the input query and fails to capture semantic similarities. Moreover, by predicting a sequence of passages each conditioned on the previously selected passages in the autoregressive ranker, the generation model gains further improvements over the strong contrastive dense ranker. We also compare Recall@k for different rankers w.r.t the \textit{oracle} ranking in Table~\ref{tab:recallk_ranker}. %(see Table~\ref{tab:recallk_ranker} in Appendix). 
The score indicates the proportion of \textit{oracle} passages (obtained y ROUGE-2 method) that is found in the top-\textit{k} predicted passages by any of the rankers. We find that autoregressive outperforms the other two rankers. %\textcolor{red}{Recall@k for different rankers w.r.t the \textit{oracle} ranking is reported in the Appendix.}

\subsection{Human Evaluation}
% \SC{One sentence on you evaluate factuality and faithfulness} % in camera ready
Here, we evaluate factuality and faithfulness of generated descriptions on AMT. 

\noindent\textbf{Factuality} ($r\xleftrightarrow{} h$). \space We evaluate the factuality of generated paragraphs using human annotators. We randomly sample 100 datapoints from the test set and evaluate paragraphs generated by BART-L using four rankers: tf-idf, contrastive dense, autoregressive and ROUGE-2 \textit{(oracle)} (a total of 400 generation examples). We ask 3 judges from AMT to evaluate the recall-oriented and precision-oriented factual correctness of each sample generation. We use the same annotation layout described for evaluating MAFE metric (correlation analysis; \S\ref{subsec:auto-eval}). More details can be found in Appendix~\ref{supp:human-design}.

Table~\ref{tab:human-eval} shows that human annotators consistently rate the factuality of paragraphs generated using autoregressive ranker higher than those generated using contrastive dense ranker and lower than \textit{Oracle} ranker. The result is consistent with our proposed metric as well. 

\begin{table}[t]
\footnotesize
    \centering
    \setlength\tabcolsep{6pt}
\renewcommand{\arraystretch}{1.5}

\begin{tabular}{lcc}
\toprule
\textbf{Ranker}   & \textbf{Recall@5} & \textbf{Recall@10} \\
\midrule
Tf-idf            & 32.02             & 42.35              \\
Contrastive Dense & 36.62             & 45.90              \\
Autoregressive    & \textbf{44.67}             & \textbf{52.08}             \\
\bottomrule
\end{tabular}

    % \vspace{-5pt}
    \caption{Recall@k (\%) for different rankers w.r.t \textit{oracle} ranking. }
    \label{tab:recallk_ranker}
\end{table}

\begin{table}[t]
\footnotesize
    \centering
    % Please add the following required packages to your document preamble:
% \usepackage{multirow}

\setlength\tabcolsep{8pt}
\renewcommand{\arraystretch}{1.6}
\begin{tabular}{lcc}
\toprule
\textbf{Ranker}             & \textbf{Recall} & \textbf{Precision} \\ \cline{1-3}
Tf-idf &             48.95              &           43.36                             \\
Contrastive Dense &              51.98              &             57.50                                \\
Autoregressive  &                \textbf{56.76}            &             \textbf{58.41}                               \\ \hdashline
\textit{Rouge2 (oracle)}   &          58.92                  &                 62.00           \\ \toprule
\end{tabular}

    \caption{Human evaluation of factuality (recall- and precision-oriented in \%) for BART-L generated paragraphs using different rankers.}
    \label{tab:human-eval}
\end{table}

\begin{table}[t]
\footnotesize
    \centering
    \setlength\tabcolsep{5.5pt}
\renewcommand{\arraystretch}{1.5}
\begin{tabular}{lccccc}
\toprule
 & \textbf{T5-Large} & \textbf{BART-L}  & \textbf{PEGASUS} \\ \hline
\textbf{Human Rating}  & \textbf{4.17} & 3.53 & 3.83  \\
\bottomrule
\end{tabular}
    % \vspace{-5pt}
    \caption{Human evaluation of faithfulness of different baselines w.r.t grounding passages. Scores are on a scale of 1 (very poor) to 5 (very high).}
    \label{tab:human-eval-faithful}
\end{table}

\noindent\textbf{Faithfulness} ($P^k \xrightarrow{}{} h$). \space We also evaluate whether the generated outputs are faithful to the top-$k$ grounding passages.\footnote{Here, we are evaluating faithfulness wrt input groundings. Thus, we use the same set of groundings (by fixing the ranker to be autoregressive) and evaluate different underlying LM.} For this, we randomly sample 100 data points from the test set and % evaluate the generated paragraphs by different baselines using human annotators from AMT. 
ask 3 annotators from AMT to evaluate the faithfulness of generated outputs using different baselines on a scale of 1-5. Following our previous annotation layout, we show one sentence at a time and then average scores across all sentences. Table~\ref{tab:human-eval-faithful} shows that T5 generates more faithful paragraphs compared to other baselines.

% \noindent\textbf{Ablations}\space 
\subsection{Ablation Studies}
Here, we discuss different ablations of our task % back in camera ready
where we remove/add certain information from/to the input and investigate its effect on the performance. We experiment with settings where there are \textit{no groundings},  \textit{no keys}, \textit{no factual keys}, \textit{no topical keys}, \textit{values w/o groundings}, and \textit{values w/ groundings}.

Table~\ref{tab:ablate} shows the results for the BART-L baseline with the autoregressive ranker. As expected, the model performance degrades the most w.r.t all metrics when the grounding passages are removed from the input. % \textcolor{blue}{This is intuitive as passages are the main source of knowledge without them} 
 This setting is similar to the prompt-to-text generation, where the model mostly relies on its parametric knowledge and is prone to hallucination.
Removing all the keys from the input is detrimental in recalling important information, as shown from the MAFE-R score.
We also observe that ablating factual keys hurts the relevance of the generated paragraph (i.e., Recall) w.r.t its reference more, whereas ablating topical keys hurts the $n$-gram overlapping metric (R-L). This is because factual keys are essential to make a good content selection and be rewarded by MAFE metric, whereas topical keys mostly appear verbatim in the output. %However, the model needs to include correct value of the factual keys to be rewarded in the PARENT score. 
Lastly, having the gold values for the corresponding keys without the grounding passages cannot beat the performance with the original inputs. In particular, although the model can recover more information (i.e. better recall), not being grounded causes it to generate less consistent information (i.e. lower precision). This is in line with our previous findings where passages play an important role in achieving good performance. % back in camera ready
% since groundings are important to achieve a narrative structure. 
When accompanied with groundings, the model achieves the best performance, emphasizing the importance of grounding.

\begin{table}[t]
\footnotesize
    \centering
    % Please add the following required packages to your document preamble:
% \usepackage[normalem]{ulem}
% \useunder{\uline}{\ul}{}
\setlength\tabcolsep{2.5pt}
\renewcommand{\arraystretch}{1.5}
\begin{tabular}{lccccc}
\hline
\toprule
\multirow{2}{*}{\textbf{Ablated Inputs}}  & \multirow{2}{*}{\textbf{R-L}} &  \multirow{2}{*}{\textbf{BERT-S}} & & \multicolumn{2}{c}{\textbf{MAFE}}               \\ \cline{5-6}          &           &           &                          &  Recall & Precision\\
% Orig. Task Input            &   8.69   &  35.80   &  16.62   &   32.97  &   59.29    &   18.47   &    25.40   &           &        &      \\ \hline   
\midrule
\rowcolor{blue!10} \multicolumn{6}{c}{\textbf{Grounding Passages}} \\
\textit{no groundings}      &   25.44   &  84.80  & &    8.87  &    13.01     \\
\midrule
\rowcolor{blue!10} \multicolumn{6}{c}{\textbf{Keys}} \\
\textit{no keys}             &   30.34     &  85.95   & &   17.99  &      27.12     \\
\textit{no factual keys}       &   \textbf{31.92}    &   \textbf{86.35}   & &   18.21  &      27.14      \\
\textit{no topical keys}     &   31.67  &    86.33  & &  \textbf{19.13}   &     \textbf{28.15}      \\ \hline
Orig. Task Input    &   32.97   &  86.58   & &   19.11  &  26.71          \\
\midrule
\rowcolor{gray!10} \multicolumn{6}{c}{\textbf{\textit{Values \& Grounding Passages}}} \\
\rowcolor{gray!10}\textit{values w/o groundings}    &  \textit{28.82}   &   \textit{85.90}  & &   \textit{19.30}   &       \textit{25.97}   \\
\rowcolor{gray!10}\textit{values w/ groundings} &   \textit{33.61}     &   \textit{86.70}  & &   \textit{21.77}   &      \textit{29.40}   \\ % \hline
% Orig. Task Input    &   32.97   &  86.58   & &   19.11  &  26.71          \\
\bottomrule
\end{tabular}

    \caption{Ablation study: Best results are in bold. The gray section is when values are assumed to be at hand, and akin to \textit{oracle} experiment.}
    \label{tab:ablate}
\end{table}

\begin{comment}
\subsection{Zero-shot and Few-shot Performance}

In this section, we experiment with our task in a zero-shot and few-shot setting. Similar to our main experiments, the input is in the form of \texttt{[Entity]} $e$ \texttt{[Title]} $t$ \texttt{[Keys]} \kfac + \ktop \texttt{[docs]} \psgs. 

\noindent\textbf{Zero-shot.} \space We choose GPT-3~\cite{NEURIPS2020_1457c0d6} as our zero-shot baseline. Given the input, we let GPT-3 to generate 200 tokens. We then stop whenever an end of sentence token was generated.
%\FB{@Michel, is it the Davinci model?} % MG: yes. 

\noindent\textbf{Few-shot.} \space We prompt GPT-3 with \textcolor{red}{X} training examples % of the form \texttt{Entity:} $e$ \texttt{Title:} $t$ \texttt{Keys:} \kfac + \ktop 
followed by each test example, and then generate the output.
\end{comment}

\section{Related Work}
\label{sec:background}

\noindent\textbf{Natural Language Generation.} \space 
Several data-to-text problems have been proposed with various input formats like Knowledge Graphs~\cite{koncel-kedziorski-etal-2019-text, cheng-etal-2020-ent}, Abstract Meaning Representations~\cite{flanigan-etal-2016-generation, amr}, tables and tree structured semantic frames~\cite{Bao2018TabletoTextDT, chen-etal-2020-logical, parikh-etal-2020-totto, chen-etal-2021-wikitablet, nan-etal-2021-dart}, and Resource Description Framework~\cite{gardent-etal-2017-webnlg}.

% The input of data-to-text generation tasks are often over-specified where some of the knowledge are already available in the parametric knowledge of recent PLMs. 
Towards a more controlled generation task, ToTTo~\cite{parikh-etal-2020-totto} was introduced for an open-domain table-to-text generation where only some of the cells are selected as the input. However, ToTTo and most existing datasets such as \textsc{WikiBio}~\cite{lebret-etal-2016-neural} and LogicNLG~\cite{chen-etal-2020-logical} focus on generating single sentences.  Although generating long-form text is becoming a new frontier for NLP research~\cite{roy-etal-2021-efficient, brahman-etal-2021-characters-tell}, not many datasets and tasks have been proposed to explore this new direction. Available datasets such as \textsc{RoToWire}~\cite{wiseman-etal-2017-challenges} or MLB~\cite{puduppully-etal-2019-data} are either small-scale or %only focus on a % back
on single domain (e.g., Sports). Unlike prior works, we propose a long-form grounded keys-to-text generation task that covers multiple domains and categories, including people, location, organization, event, etc. 

Recently, \citet{chen-etal-2021-wikitablet} presented the \textsc{WikiTableT} dataset for long-form text generation from multiple tables and meta data. However, this setting is overpecified because knowledge about entities may not always be available in structured format and may get updated in real-time. In a more natural setting, our \wgen dataset uses factual and topical keys as guidance but still leaves a considerable amount of content selection from grounding passages to be done by the model. 

%\FB{Thinking of removing this para for this submission to save space?}
There has been several work on open-ended NLG (e.g., prompt-to-text or outline-to-text)~\cite{fan-etal-2018-hierarchical, xu-etal-2018-skeleton, Yao:19, rashkin-etal-2020-plotmachines, brahman-etal-2020-cue}. Our task is also closely related to query-focused multi-document summarization~\cite{xu-lapata-2020-coarse,10.1162/tacl_a_00480} which relies on retrieval-style methods for estimating the relevance between queries and text. Additionally, our task setup can benefit from evaluation methods in summarization domain.

\noindent\textbf{Factual Consistency Evaluation.}\space\space %Despite several shortcomings of string overlap-based metrics such as BLEU and ROUGE, their simplicity has made them one of the standard and de facto automatic measures for many natural language generation tasks. 
Evaluating factual consistency of machine-generated outputs has gained growing attention in recent years. 
New approaches have been proposed mainly for tasks like abstractive summarization and machine translation~\cite{bert-score,sellam-etal-2020-bleurt, durmus-etal-2020-feqa}. Some of these metrics are QA based and have been used to measure common information between documents/reference and summaries~\cite{eyal-etal-2019-question, scialom-etal-2019-answers, wang-etal-2020-asking}. Our proposed metric, MAFE, is inspired by these works. % Inspired by these, we propose a metric for evaluating factual consistency of grounded text generation.

\section{Conclusion}
%\section{Conclusions and Future Works}

We present a practical task of grounded \textit{keys-to-text} generation and construct a large-scale dataset \wgen \space to facilitate research on this task. Experiments show the effectiveness of the proposed rankers to fetch relevant information required to generate a factual description. The human evaluation shows that \wgen \space poses a challenge to state-of-the-art models in terms of achieving human-level factuality in long-form generation.
% Models trained on our dataset can have potential applications in assistive tools for human authors, or personalized education. ...
%
Our proposed dataset and task can also foster further research in the recently emerging retrieval augmented generations models~\cite{NEURIPS2020_6b493230, zhang2021joint, Shuster2021RetrievalAR} -- where the retriever and generator components are trained end-to-end.

\section*{Limitations}
\label{app:limitation}

One of the limitations of our work is the reliance on a strong retriever/ranker. A weak retriever may result in generating text that are less factual and thus less thrust-worthy. While we proposed efficient and simple methods for training the retriever, these require large GPUs.
Additionally, as the retrieved passages get longer the quality of text generation may degrades due to known issues with encoding longer sequences. 

\section*{Acknowledgments}
We thanks our anonymous reviewers, Felix Faltings, and members of the DL and NLP group at Microsoft Research for their constructive feedback. This work was supported in part by NSF grant IIS-2047232.
% \textbf{Do not include this section when submitting your paper for review.}

% \section{Ethics Statement}

\bibliographystyle{acl_natbib}
\bibliography{acl2021}
%\bibliography{anthology,acl2021} %causes duplicates, and slow

\appendix

\clearpage
\appendix
\section{Implementation Details}
\label{app:imp_detail}

\noindent\textbf{Baselines. } Top-$10$ grounding passages were used to train and test all baselines.
We use the Transformer library~\cite{Wolf2019HuggingFacesTS}. Each baseline was trained for 3 epochs with effective batch size of 8, and initial learning rate of 5e-6 for T5 and BART, and 1e-4 for PEGASUS. We use the maximum input length of 512 tokens. During inference, we use beam search decoding with 5 beams, and repetition penalty of 1.2. Note that we use the BART-L model finetuned on XSUM dataset as our initial weights. Similarly, we use google's PEGASUS model finetuned on XSUM. The experiments are conducted in PyTorch framework using Quadro RTX 6000 GPU.

\noindent\textbf{Rankers. } The contrastive dense ranker was trained for 10 epochs with 2e-4 learning rate. The autoregressive ranker was trained for total of 30,000 steps with learning rate and weight decay of 1e-5 and 0.01, respectively. Rankers were trained using 4x Nvidia V100 GPU machines, each with 32G memory.

\noindent\textbf{Question Generation in MAFE. } The question generation module (QG) in MAFE evaluation metric, generates questions using beam search decoding with beam size of 10.

\begin{figure}[t!]
\centering
\includegraphics[width=0.9\linewidth]{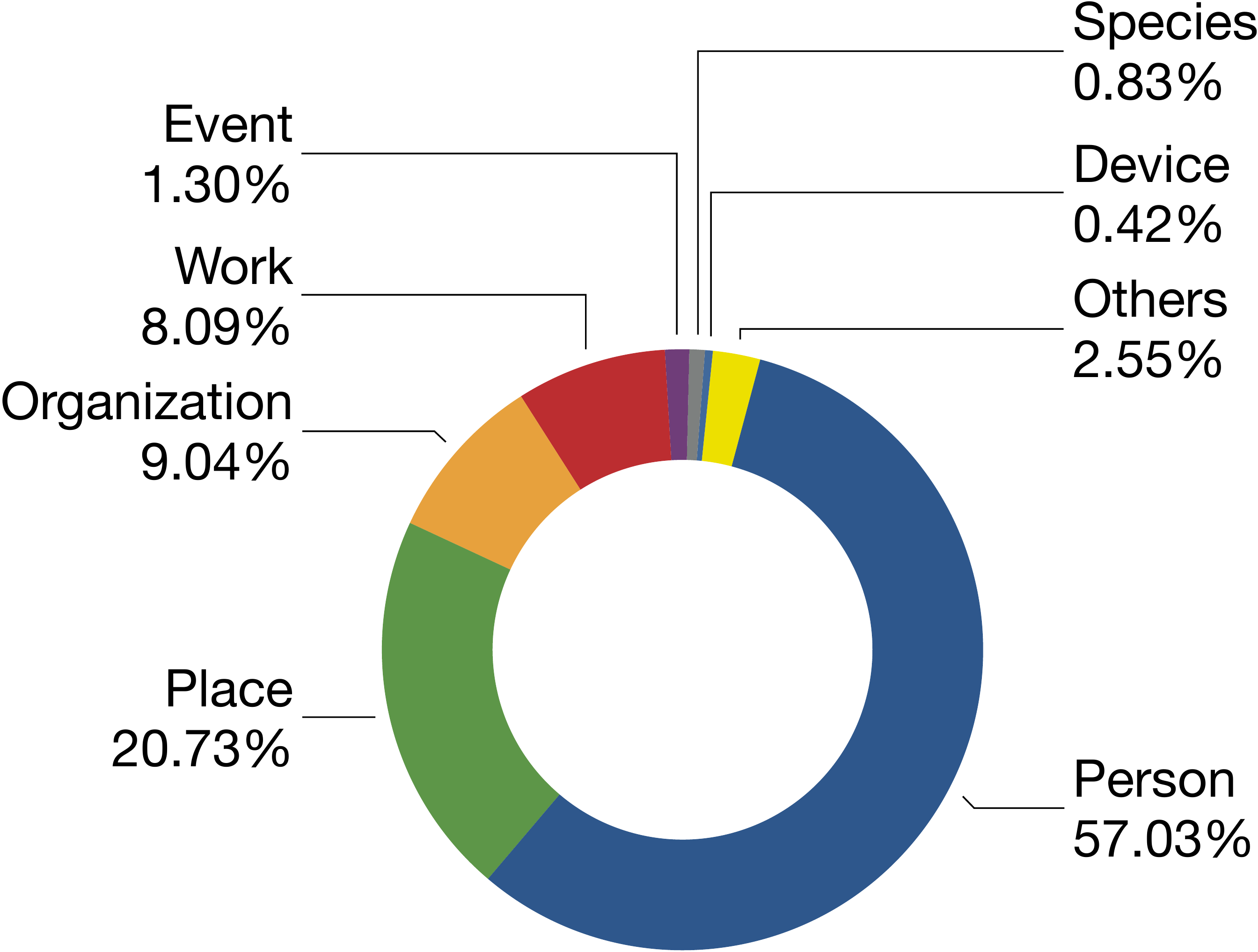}
% \vspace{-5pt}
\caption{\wgen Entity Domain Distribution.}
\label{fig:ent_type}
\end{figure}

\section{Dataset Quality Assessment}
\label{app:quality-assess}
We conducted a human evaluation on Amazon Mechanical Turk to assess the quality of our automatically constructed dataset. In this experiment, we randomly sample 100 examples from the test set. For each example, we ask 3 annotators to read the reference description carefully and answer whether each of the factual key and value pair is stated in the description or can be implied by the description. We then take the majority vote between the annotations. The result shows that 74\% of reference descriptions contain information about more than half of the key-value pairs, with Fleiss’ Kappa of 0.53 showing moderate agreement.

\section{Experimental Results}
\subsection{Automatic Evaluation}
We report all ROUGE-1 (unigram overlap), ROUGE-2 (bigram overlap) and ROUGE-L (longest matching sequence) scores in Table~\ref{tab:rouge-all}.

\begin{table}[t]
\footnotesize
    \centering
    %%% earlier version with R-1 R-2 and Par-p, and Par-R 
\setlength\tabcolsep{5pt}
\renewcommand{\arraystretch}{1.5}
\begin{tabular}{l|l|ccc}
\hline
\textbf{Model}                     & \textbf{Ranker}        & \textbf{R-1} & \textbf{R-2} & \textbf{R-L} \\ \hline
%\textbf{Extractive-QA}
\textbf{Extractive}
& n/a        &      22.37    &     6.28     &    19.85    \\ \hline
\multirow{5}{*}{\textbf{T5-Large}} & Unranked                &        28.74       &       10.31       &       26.8     \\
                                  & Tf-idf    &       29.49  &   10.54  &  26.70      \\
                                  & Contrastive Dense            &   30.92  &  12.19   &  28.14     \\
                                  & Autoregressive          &   31.45  &  12.59   &   28.64   \\ \cdashline{2-5} 
                                  & \textit{Rouge2 (oracle)} &    33.75   &   14.84  &   30.84       \\ \hline
\multirow{5}{*}{\textbf{BART-L}}   & Unranked                &      33.52   &  14.39  &   30.82      \\
                                  & Tf-idf                 &  34.00   &   14.54  &  31.14    \\
                                  & Contrastive Dense     &  35.26   &  16.04   &   32.46   \\
                                  & Autoregressive          &    \textbf{35.80}  &  \textbf{16.62}   &   \textbf{32.97} \\ \cdashline{2-5} 
                                  & \textit{Rouge2 (oracle)} &      37.72   &   18.57  &  34.87   \\ \hline
\multirow{5}{*}{\textbf{PEGASUS}}  & Unranked       &  29.23   &  12.46   &   27.11    \\
                                  & Tf-idf      &       29.43       &      12.38        &       27.25   \\
                                  & Contrastive Dense      &  31.21   &  14.16   &   28.94     \\
                                  & Autoregressive       &  32.03   &  14.90   &  29.75    \\ \cdashline{2-5} 
                                  & \textit{Rouge2 (oracle)} &     34.01  &  17.03   &   31.71  \\ \hline
\end{tabular}
\setlength\tabcolsep{8pt}
    % \vspace{-5pt}
    \caption{ROUGE scores \cite{lin2004rouge}. }
    %\caption{Results (ROUGE; \cite{lin2004rouge}). }
    \label{tab:rouge-all}
\end{table}

% s

% \begin{table}[t]
% \footnotesize
%     \centering
%     \input{tables/recallk_ranker}
%     % \vspace{-5pt}
%     \caption{Recall@k (\%) for different rankers w.r.t \textit{oracle} ranking. }
%     \label{tab:recallk_ranker}
% \end{table}

\subsection{Human Evaluations}
\label{supp:human-design}

For all the human evaluations, we restricted the pool of workers to those who were located in the US, or CA, and had a $95\%$ approval rate for at least $1,000$ previous annotations. Additionally, to further ensure the quality of annotations, we only hired master turkers, i.e., high performing turkers who have demonstrated excellence across a wide range of tasks and are awarded Masters Qualification. We also designed our setup to avoid annotator fatigue by asking them to read each paragraph only once and continuously answer several questions about it. We use a pay rate of \$15 per hour approximately based on our estimation of time needed to complete the task.

\begin{figure*}[t]
\centering
\includegraphics[width=1.0\linewidth]{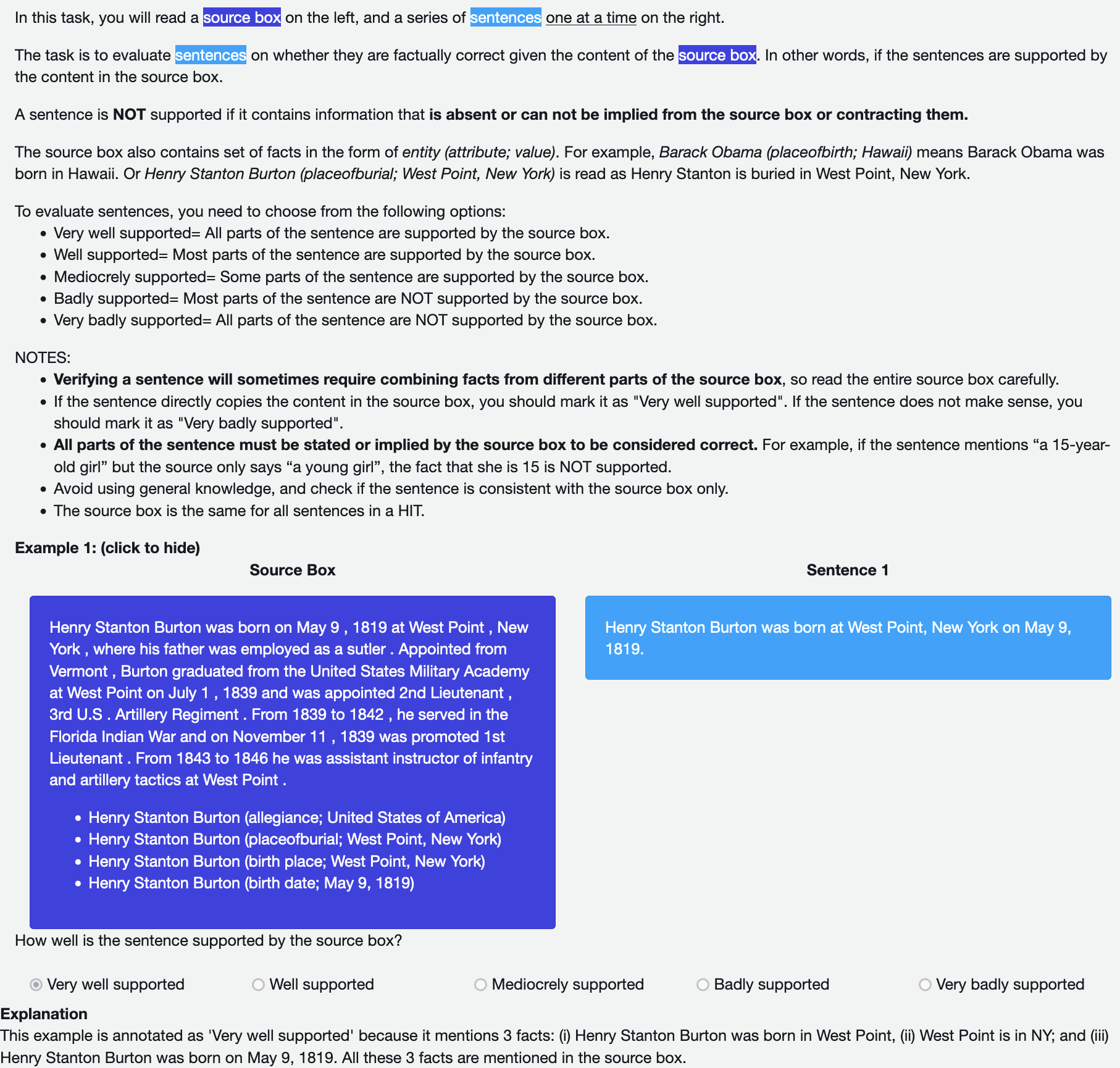}
\caption{An illustration of human evaluation of precision-oriented factuality. Generated paragraphs are presented one sentence at a time and are evaluated on how well they are supported by the references.}
\label{fig:ss-prec}
\end{figure*}

\begin{figure*}[t]
\centering
\includegraphics[width=1.0\linewidth]{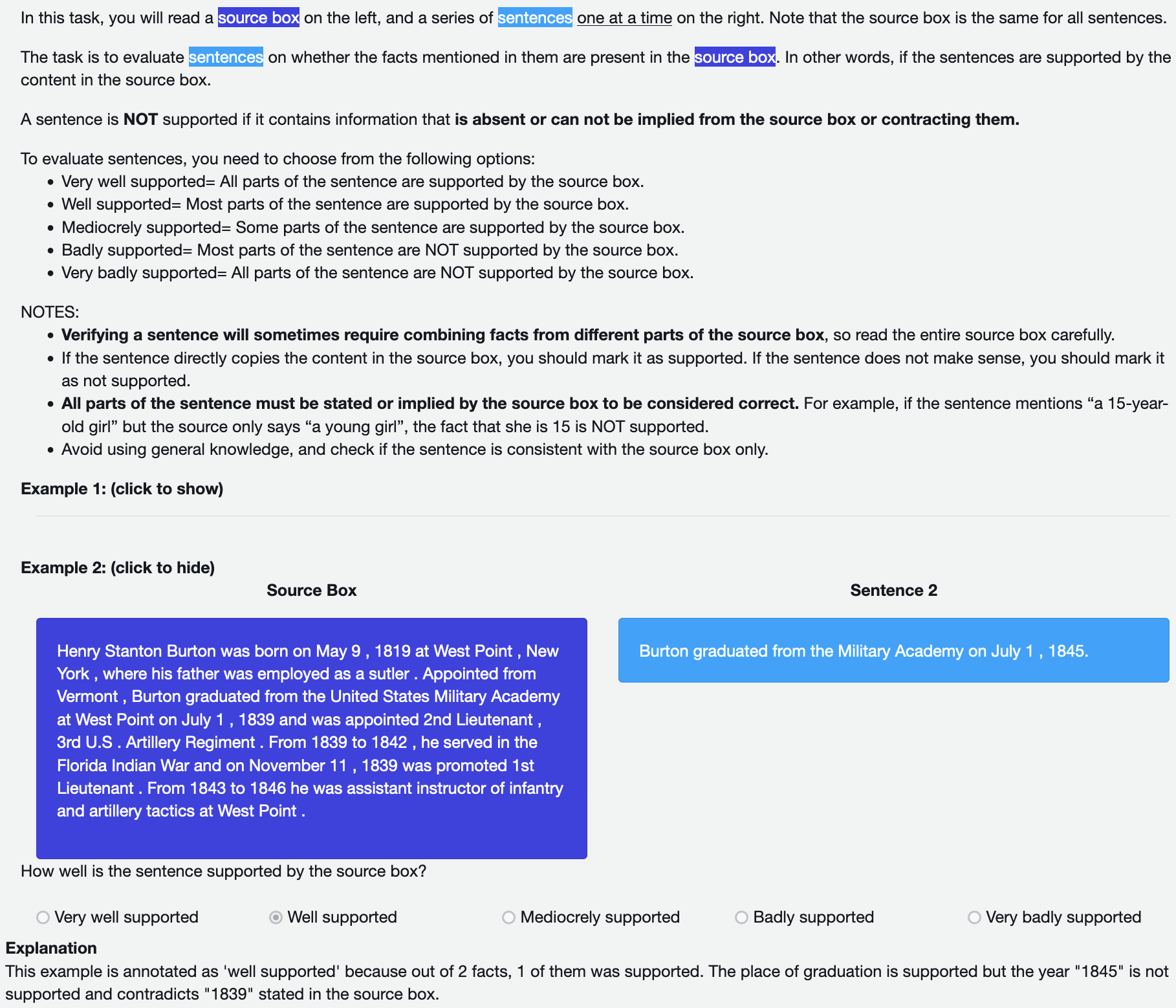}
\caption{An illustration of human evaluation of recall-oriented factuality w.r.t reference. References are presented one sentence at a time and are evaluated on how well they are supported by the generated paragraphs.}
\label{fig:ss-rec-ref}
\end{figure*}

\begin{figure*}[t]
\centering
\includegraphics[width=1.0\linewidth]{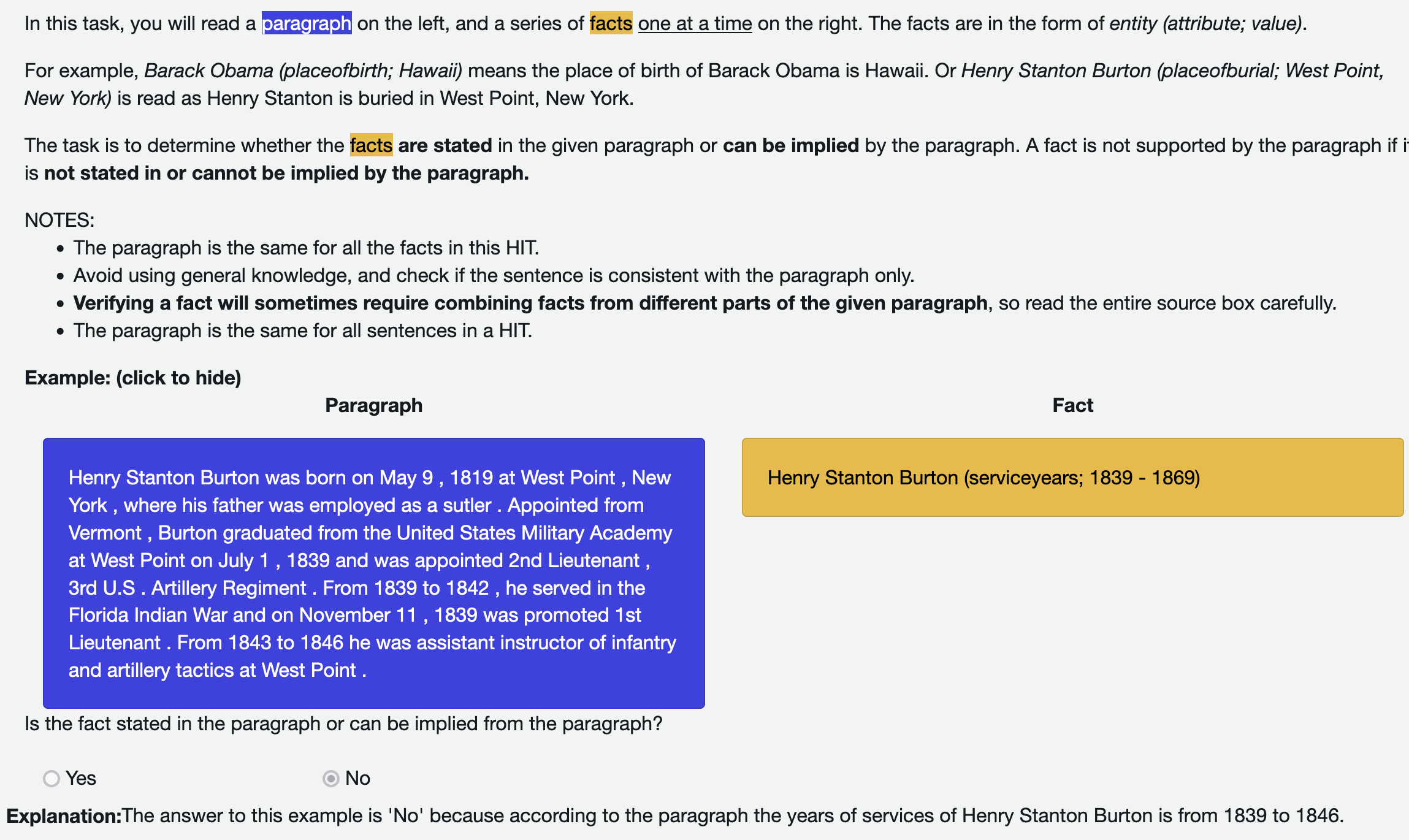}
\caption{An illustration of human evaluation of recall-oriented factuality w.r.t factual triples. Factual triples are presented one at a time and are evaluated on whether they are supported by the generated paragraphs or not.}
\label{fig:ss-rec-tr}
\end{figure*}

We depict our annotation layouts for evaluating precision-oriented and recall-oriented (both w.r.t reference and factual triples) factuality in Figure~\ref{fig:ss-prec},~\ref{fig:ss-rec-ref}, and \ref{fig:ss-rec-tr}. Likert scale of 1-5 and binary scores (supported/not supported) are used when evaluating recall w.r.t references, and factual triples, respectively. These scores are then normalized and averaged to obtain the final recall-oriented score.
% When evaluating recall w.r.t references, human annotators rate them on scale of 1-5, whereas evaluating recall w.r.t factual triples are
% Final recall score is reported by first normalizing scores obtained by 

\begin{figure}[t!]
\centering
\includegraphics[width=0.5\linewidth]{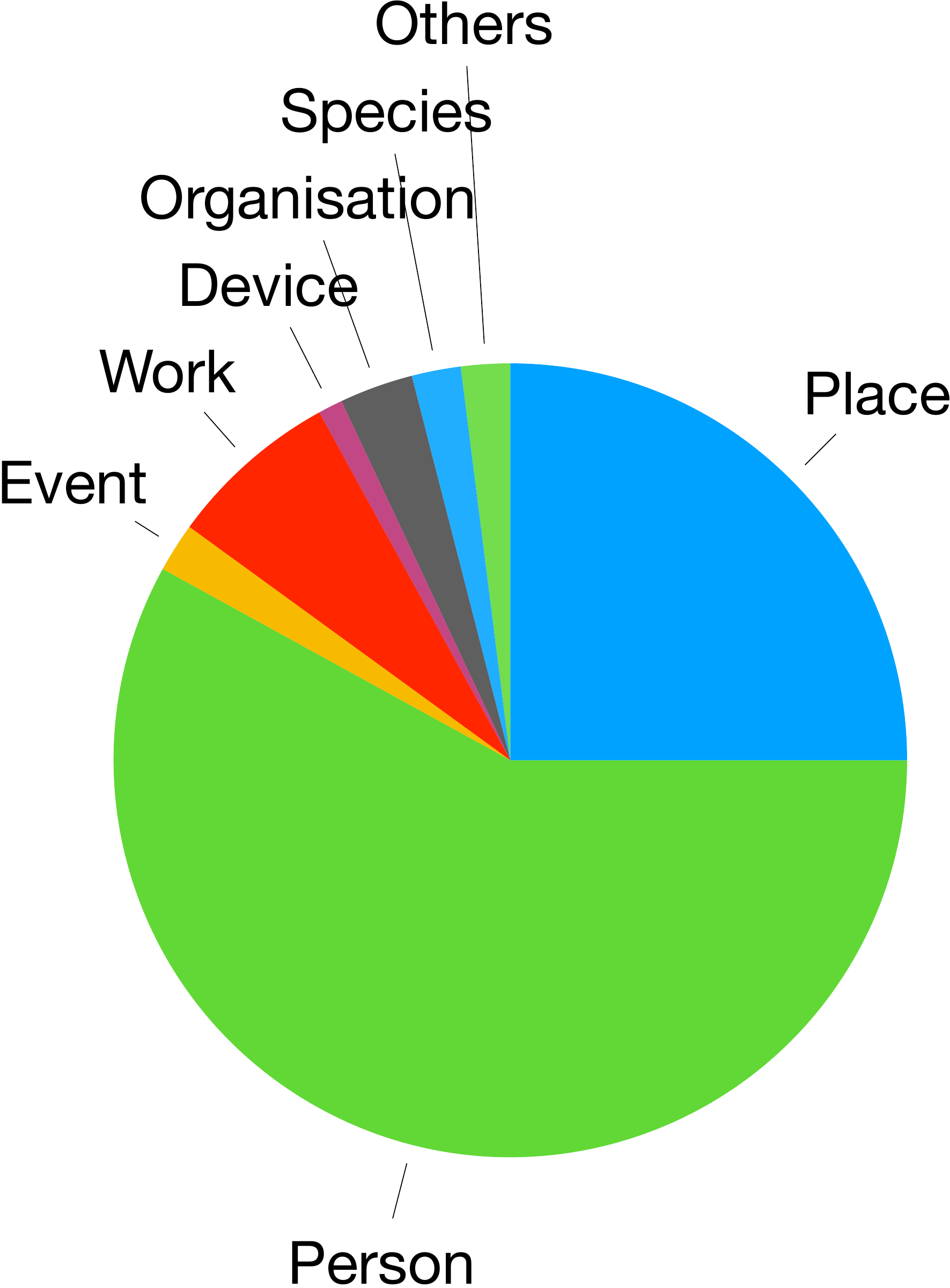}
% \vspace{-5pt}
\caption{\wgen Entity Domain Distribution of Correlation Analysis Subset.}
\label{fig:ent_type_corr}
\end{figure}

\begin{table}[!ht]
\setlength\tabcolsep{5pt}
\renewcommand{\arraystretch}{1.5}
\footnotesize
\begin{tabular}{lcc}
\hline
\textbf{Answers}                                & \textbf{F1}                    & \textbf{NLI}                  \\ \hline
gold: "saxophone"                      & \multirow{2}{*}{0.0}  & \multirow{2}{*}{1.0}  \\
predicted: "saxophonist"               &                       &                       \\ \hline
gold:"an american lawyer"              & \multirow{2}{*}{0.66} & \multirow{2}{*}{0.0}  \\
predcited:"an american politician"     &                       &                       \\ \hline
gold: "st frideswide 's priory"        & \multirow{2}{*}{0.75} & \multirow{2}{*}{1.0}  \\
predicted: "priory of st frideswide"   &                       &                       \\ \hline
gold: "december 30 , 1995"             & \multirow{2}{*}{0.75} & \multirow{2}{*}{0.0}  \\
predicted: "december 31 , 1995"        &                       &                       \\ \hline
gold: "the united kingdom"             & \multirow{2}{*}{0.8}  & \multirow{2}{*}{1.0}  \\
predicted: "united kingdom"            &                       &                       \\ \hline
gold: "his son, malcom"                & \multirow{2}{*}{0.4}  & \multirow{2}{*}{1.0}  \\
predicted: "malcom"                    &                       &                       \\ \hline
gold: "species survival plans"         & \multirow{2}{*}{0.0}  & \multirow{2}{*}{0.89} \\
predicted: "captive breeding programs" &                       &                       \\ \hline
gold: "rio de janeiro"                 & \multirow{2}{*}{0.74} & \multirow{2}{*}{1.0}  \\
predicted: "rio de janeiro , brazil"   &                       &                       \\ \hline
gold: "liberal party"                  & \multirow{2}{*}{0.5}  & \multirow{2}{*}{0.0}  \\
predicted: "conservative party"        &                       &                       \\ \hline

\end{tabular}
\caption{Examples of comparison between F1 and NLI scores. }
\label{tab:app:f1-nli}
\end{table}

% \newpage
% \bibliographystyle{acl_natbib}
% \bibliography{anthology,acl2021}
%\end{document}

\end{document}